\pdfoutput=1

\documentclass[11pt,a4paper]{article}
\usepackage[final]{acl2022}
\usepackage{times}
\usepackage{latexsym}

\usepackage[utf8]{inputenc} 
\usepackage[T1]{fontenc}    
\usepackage{hyperref}       
\usepackage{url}            
\usepackage{booktabs}       
\usepackage{graphicx}
\usepackage{amsfonts}       
\usepackage{nicefrac}       
\usepackage{microtype}      
\usepackage{CJKutf8}
\usepackage{amsmath}
\usepackage[linesnumbered,ruled,vlined]{algorithm2e}
\usepackage{multicol}
\usepackage{multirow}
\usepackage{arydshln}
\usepackage{paralist}
\usepackage[american]{babel}
\usepackage{microtype}
\usepackage{hhline}

\title{Are Pre-trained Transformers Robust in Intent Classification?\\A Missing Ingredient in Evaluation of Out-of-Scope Intent Detection}

\author{
  \textbf{Jianguo Zhang}$^1$~~~~ \textbf{Kazuma Hashimoto}$^2$~~~~  \textbf{Yao Wan}$^3$~~~~ \textbf{Zhiwei Liu}$^4$ \\  \textbf{Ye Liu}$^1$~~~~ \textbf{Caiming Xiong}$^1$~~~~   \textbf{Philip S. Yu}$^4$  \\
  $^1$Salesforce Research, Palo Alto, USA \\
 $^2$Google Research, Mountain View, USA\\
   $^3$Huazhong University of Science and Technology, Wuhan, China \\
    $^4$University of Illinois at Chicago, Chicago, USA \\ 
{\texttt{jianguozhang@salesforce.com}}
}

\begin{document}
\maketitle
\begin{abstract}
Pre-trained Transformer-based models were reported to be robust in intent classification. In this work, we first point out the importance of in-domain out-of-scope detection in few-shot intent recognition tasks and then illustrate the vulnerability of pre-trained Transformer-based models against samples that are in-domain but out-of-scope (ID-OOS). We construct two new datasets, and empirically show that pre-trained models do not perform well on both ID-OOS examples and general out-of-scope examples, especially on fine-grained few-shot intent detection tasks.   
To figure out how the models mistakenly classify ID-OOS intents as in-scope intents, we further conduct analysis on confidence scores and the overlapping keywords, as well as point out several prospective directions for future work. Resources are available at \url{https://github.com/jianguoz/Few-Shot-Intent-Detection}.

\end{abstract}

\section{Introduction}
\label{sec:intro}


Intent detection, which aims to identify intents from user utterances, is a vital task in goal-oriented dialog systems~\citep{xie2022converse}.
However, the performance of intent detection has been hindered by the data scarcity issue, as it is non-trivial to collect sufficient examples for new intents. 
In practice, the user requests could also be not expected or supported by the tested dialog system, referred to as out-of-scope (OOS) intents.
Thus, it is important to improve OOS intents detection performance while keeping the accuracy of detecting in-scope intents in the few-shot learning scenario.

Recently, several approaches~\citep{ood_intent,zhang2020discriminative,tod-bert,cavalin2020improving,zhan2021out,xu2021unsupervised} have been proposed to improve the performance of identifying in-scope and OOS intents in few-shot scenarios. 
Previous experiments have shown that a simple confidence-based out-of-distribution detection method~\cite{softmax_conf,acl-ood} equipped with pre-trained BERT can improve OOS detection accuracy.
However, there is a lack of further study of pre-trained Transformers on few-shot fine-grained OOS detection where the OOS intents are more relevant to the in-scope intents. Besides, those studies mainly focus on the CLINC dataset~\citep{oos-intent}, in which the OOS examples are designed such that they do {\it not} belong to any of the known intent classes. Their distribution is dissimilar to each other, and thus they are easy to be distinguished from the known intent classes. Moreover, CLINC is not enough to study more challenging few-shot fine-grained OOS detection as it lacks such semantically similar OOS examples to in-scope intents, and other popular used datasets such as BANKING77~\cite{casanueva2020efficient} do not contain OOS examples. 

In this paper, we 
aim to investigate the following research question: ``\textit{Are pre-trained Transformers robust in intent classification w.r.t. general and relevant OOS examples?}''.
We first define two types of OOS intents: out-of-domain OOS ({\bf OOD-OOS}) and in-domain OOS ({\bf ID-OOS}). We then investigate \textit{how} robustly state-of-the-art pre-trained Transformers perform on these two OOS types.
The OOD-OOS is identical to the OOS in the CLINC dataset, where the OOS and in-scope intents (e.g., requesting an online TV show service in a banking system) are topically rarely overlapped.
We construct an ID-OOS set for a domain, by separating semantically-related intents from the in-scope intents (e.g., requesting a banking service that is not supported by the banking system).  

Empirically, we evaluate several pre-trained Transformers (e.g., BERT~\cite{bert}, RoBERTa~\cite{roberta}, ALBERT~\cite{albert}, and ELECTRA~\cite{electra}) in the few-shot learning scenario, as well as pre-trained ToD-BERT~\cite{tod-bert} on task-oriented dialog system.
The contributions of this paper are two-fold.
First, we constructed and released two new datasets for OOS intent detection based on the single-domain CLINC dataset and the large fine-grained BANKING77 dataset. 
Second, we reveal several interesting findings through experimental results and analysis:
1) the pre-trained models are much less robust on ID-OOS than on the in-scope and OOD-OOS examples; 2) both ID-OOS and OOD-OOS detections are not well tackled and require further explorations on the scenario of fine-grained few-shot intent detection; 
and 3) it is surprising that pre-trained models can predict undesirably confident scores even when masking keywords shared among confusing intents.

\section{Evaluation Protocol}
\label{sec:methodology}


\paragraph{Task definition}
We consider a few-shot intent detection system that handles pre-defined $K$ in-scope intents.
The task is, given a user utterance text $u$, to classify $u$ into one of the $K$ classes or to recognize $u$ as OOS (i.e., OOS detection).
To evaluate the system, 
we adopt
in-scope accuracy $A_\mathrm{in} = C_\mathrm{in}/N_\mathrm{in}$, and OOS recall $R_\mathrm{oos} = C_\mathrm{oos}/N_\mathrm{oos}$, following \newcite{oos-intent} and \newcite{zhang2020discriminative}.
We additionally report OOS precision, $P_\mathrm{oos} = C_\mathrm{oos}/N'_\mathrm{oos}$.
$C_\mathrm{in}$ and $C_\mathrm{oos}$ are the number of correctly predicted in-scope and out-of-scope examples, respectively; $N_\mathrm{in}$ and $N_\mathrm{oos}$ are the total number of the in-scope and out-of-scope examples evaluated, respectively; if an in-scope example is predicted as OOS, it is counted as wrong.
$N'_\mathrm{oos}$ ($\leq N_\mathrm{in} + N_\mathrm{oos}$) is the number of examples predicted as OOS.

\paragraph{Inference}
We use a confidence-based method~\cite{acl-ood} to evaluate the five pre-trained Transformers.
We compute a hidden vector $h=\mathrm{Encoder}(u)\in\mathbb{R}^{768}$ for $u$, where $\mathrm{Encoder}\in$ \{BERT, RoBERTa, ALBERT, ELECTRA, ToD-BERT\},
and compute a probability vector $p(y|u) = \mathrm{softmax}(Wh+b)\in\mathbb{R}^{K}$, where $W$ and $b$ are the model parameters. We first take the class $c$ with the largest value of $p(y=c|u)$, then output $c$ if $p(y=c|u) > \delta$, where $\delta\in[0.0, 1.0]$ is a threshold value, and otherwise we output OOS.  $\delta$ is tuned by using the development set, so as to maximize $(A_\mathrm{in}+R_\mathrm{oos})$ averaged across different runs~\citep{zhang2020discriminative}.

\paragraph{Training}
To train the model, we use training examples of the in-scope intents, without using any OOS  examples.
This is reasonable as it is nontrivial to collect sufficient OOS data to model the large space and distribution of the unpredictable OOS intents~\cite{zhang2020discriminative,cavalin2020improving}.

\begin{table*}[h]
  \begin{center}
{
\resizebox{\linewidth}{!}{
    \begin{tabular}{l|l|l}
    \hline
    Domain & IN-OOS & In-scope \\ \hline
    Banking      & balance, bill\_due, min\_payment, & account\_blocked, bill\_balance, interest\_rate, order\_checks, pay\_bill, \\
                 & freeze\_account, transfer        & pin\_change, report\_fraud, routing, spending\_history, transactions \\ \cdashline{1-3}

    Credit  & report\_lost\_card, improve\_credit\_score, & credit\_score, credit\_limit, new\_card, card\_declined, international\_fees, \\
    cards    & rewards\_balance, application\_status,    & apr, redeem\_rewards, credit\_limit change, damaged\_card \\
            & replacement\_card\_duration               & expiration\_date \\ 

                 \hline

    \end{tabular}}
}
    \caption{Data split of the ID-OOS and in-scope intents for the CLINC dataset. }
    \label{tb:intent_split-stats}
    \vspace{-1em}
  \end{center}

\end{table*}


\begin{table*}[]
\centering
\resizebox{\linewidth}{!}{
\begin{tabular}{l|l}
\hline
ID-OOS&
    \begin{tabular}[c]{@{}l@{}}
    ``pin\_blocked'', ``top\_up\_by\_cash\_or\_cheque'' ``top\_up\_by\_card\_charge'', ``verify\_source\_of\_funds'',\\ ``transfer\_into\_account'', ``exchange\_rate'', ``card\_delivery\_estimate'', ``card\_not\_working'',\\ ``top\_up\_by\_bank\_transfer\_charge'', ``age\_limit'', ``terminate\_account'', ``get\_physical\_card'', \\
    ``passcode\_forgotten'', ``verify\_my\_identity'', ``topping\_up\_by\_card'', ``unable\_to\_verify\_identity'',\\ 
    ``getting\_virtual\_card'', ``top\_up\_limits'', ``get\_disposable\_virtual\_card'', ``receiving\_money'',\\ ``atm\_support'',
    ``compromised\_card'', ``lost\_or\_stolen\_card'', ``card\_swallowed'', ``card\_acceptance'',\\ ``virtual\_card\_not\_working'',
    ``contactless\_not\_working''
    \end{tabular}\\
\hline
\end{tabular}}
\caption{Data split of the ID-OOS intents for the BANKING77 dataset. Where 27 intents are randomly selected as ID-OOS intents and the rest are treated as in-scope intents. Here we show the 27 selected ID-OOS intents.} \label{tb:intent_split_banking77-stats}
\end{table*}

\section{Dataset Construction}

We describe the two types of OOS (i.e., OOD-OOS and ID-OOS), using the CLINC dataset~\citep{oos-intent} and the fine-grained BANKING77 dataset~\citep{casanueva2020efficient}.
The CLINC dataset covers 15 intent classes for each of the 10 different domains, and it also includes OOS examples. We randomly select two domains, i.e., the ``Banking'' and ``Credit cards'',  out of the ten domains for models evaluation.
The BANKING77 dataset is a large fine-grained single banking domain intent dataset with  77 intents, and it initially does not include OOS examples.
We use these two datasets since CLINC dataset focuses on the OOS detection task, and we can evaluate models on the large single fine-grained banking domain on BANKING77 dataset.

\paragraph{OOD-OOS}
We use the initially provided OOS examples of CLINC dataset as OOD-OOS examples for both datasets.
To justify our hypothesis that the CLINC's OOS examples can be considered as out of domains, we take 100 OOS examples from the development set, and check whether the examples are related to each domain.
Consequently, only 4 examples are relevant to ``Banking'', while none of them is related to ``Credit cards''.
There are also no overlaps between the added OOS examples and the original BANKING77 dataset.
These findings show that most of the OOS examples are not related to the targeted domains, and we cannot effectively evaluate the model's capability to detect OOS intents within the same domain.

\paragraph{ID-OOS}

Detecting the OOD-OOS examples is important in practice, but we focus more on how the model behaves on ID-OOS examples.
For the ID-OOS detection evaluation, we separate 5 intents from the 15 intents in each of the domains and use them as the ID-OOS samples for the CLINC dataset, following the previous work~\cite{open-class}. 
In contrast to the previous work that randomly splits datasets, we intentionally design a confusing setting for each domain.
More specifically, we select 5 intents that are semantically similar to some of the 10 remaining intents. 
As for the BANKING77 dataset, we randomly separate 27 intents from the 77 intents and use them as the ID-OOS samples, following the above process. 

Table~\ref{tb:intent_split-stats} and Table~\ref{tb:intent_split_banking77-stats} show which intent labels are treated as ID-OOS for the CLINC dataset and BANKING77 dataset, respectively.


\paragraph{Data Statistics}


For each domain, the original CLINC dataset has 100, 20, and 30 examples for each in-scope intent, and 100, 100, and 1000 OOD-OOS examples for the train, development, and test sets, respectively.
We reorganize the original dataset to incorporate the ID-OOS intents and construct new balanced datasets.
For each in-scope intent in the training set, we keep 50 examples as a new training set, and move the rest 30 examples and 20 examples to the development and test sets through random sampling.
For the examples of each ID-OOS intent in the training set, we randomly sample 60 examples, add them to the development set, and add the rest of the 40 examples to the test set.
We move the unused OOD-OOS examples of the training set to the validation set and keep the OOD-OOS test set unchanged. 
For the BANKING77 dataset, we move the training/validation/test examples of the selected 27 intents to the ID-OOS training/validation/test examples, and we copy the OOD-OOS examples of CLINC as the OOD-OOS examples of BANKING77. 

We name the two new datasets as CLINC-Single-Domain-OOS and BANKING77-OOS, respectively. 
Table \ref{tb:dataset} shows the dataset statistics.

\begin{table}[t!]
\centering
\resizebox{1.0\linewidth}{!}{
	\renewcommand{\arraystretch}{0.8}
\begin{tabular}{l|cccc}
\hline
\textbf{CLINC-Single-Domain-OOS} & K  & Train & Dev. & Test \\ \hline
In-scope                  & 10 & 500   & 500  & 500  \\
ID-OOS                    & -  & -     & 400  & 350  \\
OOD-OOS                   & -  & -     & 200  & 1000 \\
\hhline{=====}
\textbf{BANKING77-OOS}                 & K  & Train & Dev. & Test \\ \hline
In-scope                  & 50 & 5905  & 1506 & 2000 \\
ID-OOS                    & -  & -  & 530  & 1080 \\
OOD-OOS                   & -  & -     & 200  & 1000 \\\hline
\end{tabular}}
\caption{
Statistics of CLINC-Single-Domain-OOS and BANKING77-OOS dataset.
}
\label{tb:dataset}
\end{table}

\begin{table*}[]
\centering
\resizebox{1.0\linewidth}{!}{
\begin{tabular}{cl|ccc|ccc|ccc}
\hline
\multicolumn{1}{l}{\textbf{}} & \textbf{} & \multicolumn{3}{c|}{\textbf{In-scope accuracy}}           & \multicolumn{3}{c|}{\textbf{OOS recall}}                   & \multicolumn{3}{c}{\textbf{OOS precision}}                 \\
\multicolumn{2}{l|}{\textbf{5-shot}}      &  \textbf{Banking} & \textbf{Credit cards} & \textbf{BANKING77-OOS} & \textbf{Banking} & \textbf{Credit cards} & \textbf{BANKING77-OOS} & \textbf{Banking} & \textbf{Credit cards} & \textbf{BANKING77-OOS} \\ \hline
\multirow{4}{*}{ID-OOS}       & ALBERT    & 54.1 $\pm$ 6.9   & 55.5 $\pm$ 8.1        &  20.3 $\pm$ 2.4 & 86.3 $\pm$ 8.1   & 75.9 $\pm$ 11.2       &  89.5 $\pm$ 1.5 & 57.9 $\pm$ 3.3   & 55.8 $\pm$ 4.3        &  39.8 $\pm$ 0.7 \\
                              & BERT      & 75.2 $\pm$ 2.9   & 74.1 $\pm$ 4.6        &  25.4 $\pm$ 3.6  & 81.8 $\pm$ 10.5  & 76.5 $\pm$ 9.7        & 90.9 $\pm$ 0.6  & 70.8 $\pm$ 2.5   & 68.1 $\pm$ 3.2        & 41.3 $\pm$ 1.4 \\
                              & ELECTRA   & 64.8 $\pm$ 4.8   & 71.0 $\pm$ 7.3        &  30.9 $\pm$ 2.3 & 89.4 $\pm$ 4.3   & 75.8 $\pm$ 6.1        & 87.5 $\pm$ 2.4  & 65.1 $\pm$ 3.0   & 67.1 $\pm$ 4.8        &  43.0 $\pm$ 0.8 \\
                              & RoBERTa   & 83.8 $\pm$ 1.7   & 64.5 $\pm$ 5.6        & 43.0 $\pm$ 2.9 & 78.4 $\pm$ 6.2   & 86.8 $\pm$ 5.4        & 83.1 $\pm$ 4.3 & 78.6 $\pm$ 1.5   & 63.3 $\pm$ 3.4        & 46.3 $\pm$ 1.9  \\ 
                              
                               & ToD-BERT   & 75.1 $\pm$ 2.3   & 67.4 $\pm$ 4.2      & 35.5 $\pm$ 1.5 & 75.8 $\pm$ 9.5   &    72.3 $\pm$ 3.4   & 82.7 $\pm$ 1.8 &  69.4 $\pm$ 3.6 & 61.3 $\pm$ 2.3     & 43.8 $\pm$ 0.1  \\ 
                              \hdashline
\multirow{4}{*}{OOD-OOS}      & ALBERT    & 63.1 $\pm$ 5.7   & 55.5 $\pm$ 8.1        & 20.3 $\pm$ 2.4  & 85.3 $\pm$ 5.4   & 92.5 $\pm$ 4.0        & 97.3 $\pm$ 2.5 & 83.4 $\pm$ 1.7   & 81.5 $\pm$ 3.1        &  39.9 $\pm$ 1.3 \\
                              & BERT      & 75.2 $\pm$ 2.9   & 74.1 $\pm$ 4.6        & 39.0 $\pm$ 3.1  & 93.4 $\pm$ 3.7   & 95.5 $\pm$ 2.7        & 94.1 $\pm$ 1.6  & 88.8 $\pm$ 1.4   & 88.4 $\pm$ 1.9        & 49.0 $\pm$ 1.8 \\
                              & ELECTRA    & 75.5 $\pm$ 4.0   & 71.0 $\pm$ 7.3        & 39.1 $\pm$ 2.7 & 87.3 $\pm$ 4.3   & 87.6 $\pm$ 4.2        & 93.1 $\pm$ 4.3 & 88.8 $\pm$ 2.1   & 87.0 $\pm$ 2.7        & 48.7 $\pm$ 1.1 \\
                              & RoBERTa   & 83.8 $\pm$ 1.7   & 81.2 $\pm$ 4.0        & 62.1 $\pm$ 2.9 & 97.0 $\pm$ 0.9   & 96.7 $\pm$ 1.4        & 93.9 $\pm$ 1.4 & 92.9 $\pm$ 0.6   & 91.4 $\pm$ 1.8        & 68.7 $\pm$ 2.2 \\
                              & ToD-BERT   & 83.0 $\pm$ 1.6   &  75.8 $\pm$ 5.0    & 52.9 $\pm$ 1.5 &  91.9 $\pm$ 1.0  &  96.7 $\pm$ 0.9     & 88.4 $\pm$ 1.7 & 92.8 $\pm$ 0.6  &  89.6 $\pm$ 2.1    & 66.0 $\pm$ 1.2 \\ 
                              \hline
\multicolumn{1}{l}{\textbf{10-shot}} & \textbf{} & \multicolumn{3}{c|}{}           & \multicolumn{3}{c|}{}                   & \multicolumn{3}{c}{}                  \\ \hline
\multirow{4}{*}{ID-OOS}       & ALBERT    & 77.8 $\pm$ 2.7   & 66.7 $\pm$ 7.8        &  27.3 $\pm$ 3.4 & 77.6 $\pm$ 13.0  & 79.8 $\pm$ 6.4        & 87.6 $\pm$ 1.3 & 72.2 $\pm$ 2.9   & 64.0 $\pm$ 4.1        & 42.4 $\pm$ 1.3  \\
                              & BERT      & 77.5 $\pm$ 1.7   & 80.3 $\pm$ 3.7        & 52.5 $\pm$ 1.7  & 87.5 $\pm$ 9.2   & 74.5 $\pm$ 6.9        & 77.3 $\pm$ 3.2  & 73.8 $\pm$ 1.7   & 73.1 $\pm$ 3.3        &   50.8 $\pm$ 1.1\\
                              & ELECTRA   & 79.5 $\pm$ 2.9   & 78.0 $\pm$ 2.5        & 40.1 $\pm$ 2.7  & 85.2 $\pm$ 9.1   & 86.5 $\pm$ 5.8        &  84.0 $\pm$ 1.7 & 75.4 $\pm$ 2.7   & 73.3 $\pm$ 2.9        & 46.1 $\pm$ 1.1   \\
                              & RoBERTa   & 76.6 $\pm$ 0.9   & 81.0 $\pm$ 5.5        & 59.7 $\pm$ 1.2  & 86.4 $\pm$ 6.3   & 83.9 $\pm$ 6.9        & 79.1 $\pm$ 1.7  & 72.7 $\pm$ 1.5   & 75.8 $\pm$ 5.2        &  55.8 $\pm$ 1.1 \\
                              & ToD-BERT   & 80.7 $\pm$ 2.5   &  80.6 $\pm$ 0.9    & 54.3 $\pm$ 1.8 &  79.5 $\pm$ 6.1  & 70.2 $\pm$ 5.9       & 76.9 $\pm$ 2.7 & 75.4 $\pm$ 1.4  & 71.9 $\pm$ 2.6      & 52.1 $\pm$ 1.2 \\ 
                              \cdashline{1-11}
\multirow{4}{*}{OOD-OOS}      & ALBERT    & 77.8 $\pm$ 2.7   & 66.7 $\pm$ 7.8        & 30.5 $\pm$ 6.5 & 90.6 $\pm$ 4.0   & 95.0 $\pm$ 3.4        &  92.7 $\pm$ 6.3 & 89.8 $\pm$ 1.0   & 85.7 $\pm$ 2.7        & 47.1 $\pm$ 1.9  \\
                              & BERT      & 77.5 $\pm$ 1.7   & 90.1 $\pm$ 1.9        & 64.2 $\pm$ 0.5 & 96.8 $\pm$ 1.2   & 91.1 $\pm$ 4.4        & 91.4 $\pm$ 3.2 & 90.0 $\pm$ 0.7   & 95.5 $\pm$ 1.1        & 68.9 $\pm$ 1.0  \\
                              & ELECTRA   & 79.5 $\pm$ 2.9   & 88.6 $\pm$ 2.1        & 40.1 $\pm$ 2.7  & 94.8 $\pm$ 1.7   & 89.1 $\pm$ 2.2        & 97.6 $\pm$ 1.0 & 90.7 $\pm$ 1.2   & 94.2 $\pm$ 1.1        &  47.9 $\pm$ 1.4 \\
                              & RoBERTa   & 89.2 $\pm$ 1.3   & 87.5 $\pm$ 3.3        & 70.3 $\pm$ 0.3 & 95.6 $\pm$ 1.0   & 94.6 $\pm$ 2.4        & 94.0 $\pm$ 0.8 & 95.4 $\pm$ 0.5   & 94.0 $\pm$ 1.4        & 73.3 $\pm$ 1.5 \\
                              & ToD-BERT   &  86.5 $\pm$ 2.6  &  86.5 $\pm$ 0.6   & 60.6 $\pm$ 1.8 & 96.0 $\pm$ 0.5   &  96.4 $\pm$ 0.5      & 94.9 $\pm$ 0.9 & 94.2 $\pm$ 1.2 &    93.7 $\pm$ 0.3 & 63.3 $\pm$ 0.9  \\ 
                              \hline
\end{tabular}}
\caption{Testing results on the ``Banking'' and ``Credit cards'' domains in CLINC-Single-Domain-OOS and BANKING77-OOS datasets. Note that as the best $\delta$ is selected based on  $(A_\mathrm{in}+R_\mathrm{oos})$, the in-scope accuracy could be different in the scenarios of OOD-OOS and ID-OOS (see Figure~\ref{fig:comparisons-among-models}).}
\label{tb:main}
\vspace{-0.5em}
\end{table*}

\begin{figure*}
	\begin{center}
    	\includegraphics[width=1.0\linewidth]{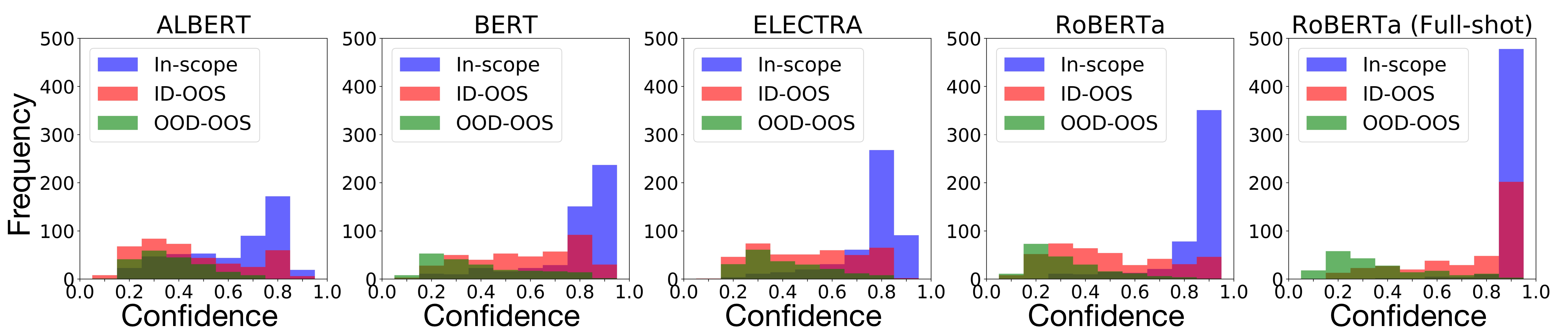}
    \end{center}
\caption{Model confidence on the development set of ``Banking'' domain in CLINC-Single-Domain-OOS dataset under 5-shot setting. Darker colors indicate overlaps.}
\label{fig:model-comp}
\vspace{-1em}
\end{figure*}

\section{Empirical Study}
\label{experiments}



\subsection{Experimental Setting}
We implement all the models following public code from \newcite{zhang2020discriminative},
based on the HuggingFace Transformers library~\citep{wolf2019huggingface} for the easy reproduction of experiments.
For each component related to the five pre-trained models, we use their base configurations. 
We use the {\tt roberta-base} configuration for RoBERTa; {\tt bert-base-uncased} for BERT; {\tt albert-base-v2} for ALBERT; {\tt electra-base-discriminator} for ELECTRA; {\tt tod-bert-jnt-v1} for ToDBERT. 
All the model parameters are updated during the fine-tuning process. 
We use the AdamW~\cite{adamwr} optimizer with a weight decay coefficient of 0.01 for all the non-bias parameters.
We use a gradient clipping technique~\cite{clip} with a clipping value of 1.0, and also use a linear warmup learning-rate scheduling with a proportion of 0.1 w.r.t. to the maximum number of training epochs. 

For each model, we perform hyper-parameters searches for learning rate values $\in \left \{ 1e-4, 2e-5, 5e-5 \right \}$, and the number of the training epochs $\in \left \{ 8, 15, 25,35 \right \}$.  We set the batch size to 10 and 50 for CLINC-
Single-Domain-OOS and BANKING77-OOS, respectively.
We take the hyper-parameter sets for each experiment and train the model ten times for each hyper-parameter set to select the best threshold $\delta$ (introduced in Section~\ref{sec:methodology}) on the development set.
We then select the best hyper-parameter set along with the corresponding threshold. Finally, we apply the best model and the threshold to the test set. Experiments were conducted on single NVIDIA Tesla V100 GPU with 32GB memory. 

We mainly conduct the experiments in 5-shot, e.g., five training examples per in-scope intent, and 10-shot; we also report partial results in the full-shot scenario.

\begin{figure*}[ht]
	\begin{center}
    	\includegraphics[width=1.0\linewidth]{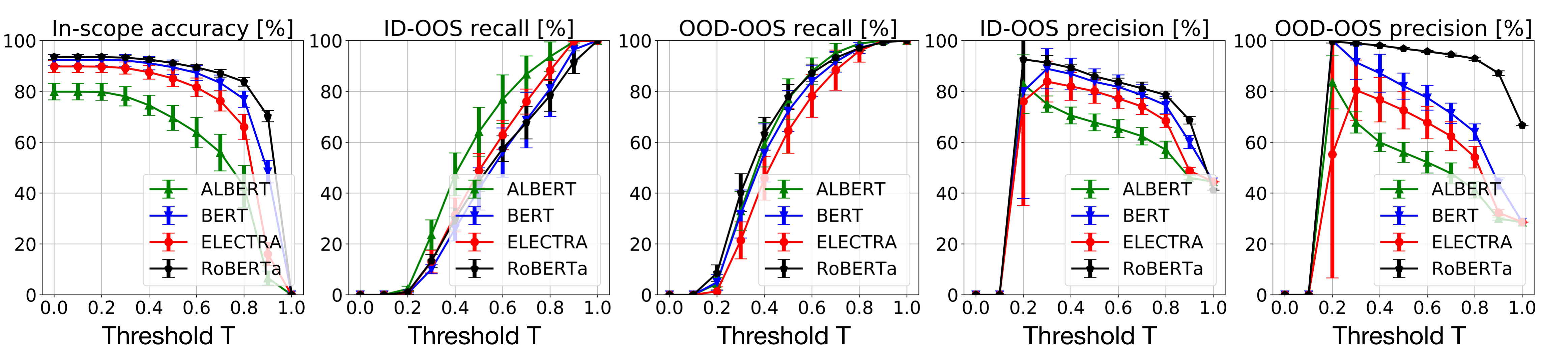}
    \end{center}
\caption{Results on the ``Banking'' domain in CLINC-Single-Domain-OOS dataset (Dev. set) under 5-shot setting.}
\label{fig:comparisons-among-models}
\vspace{-0.5em}
\end{figure*}


\begin{figure}
	\begin{center}
    	\includegraphics[width=\linewidth]{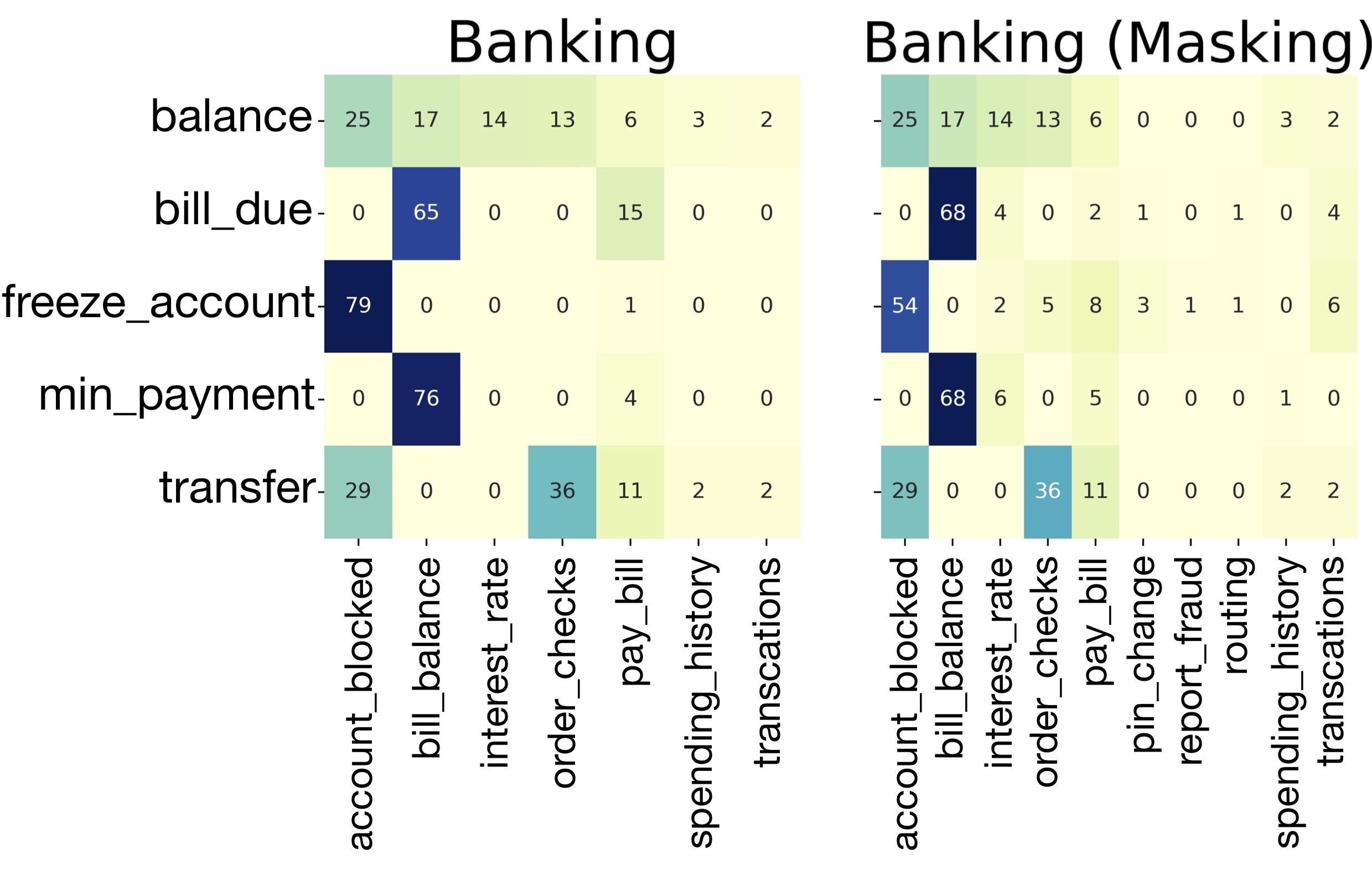}
    \end{center}
\caption{Full-shot confusion matrices on the development set with and without masking (``Banking'', RoBERTa). Vertical axis: ID-OOS; horizontal axis: in-scope (only predicted intents considered). 
}
\label{fig:conf_mat}
\vspace{-1.3em}
\end{figure}

\subsection{Overall Results}
\label{subsec:overall}

Table~\ref{tb:main} shows the results of few-shot intent detection on the test set for 5-shot and 10-shot settings.
In both settings, the in-scope accuracy of ID-OOS examples tends to be lower than that of OOD-OOS examples, and the gap becomes larger for OOS recall and precision. It is interesting to see that ToD-BERT, which is pre-trained on several task-oriented dialog datasets, does not perform well in our scenario. The results indicate that the pre-trained models are much less robust on the ID-OOS intent detection. Compared with the results on the two single domains of the CLINC-Single-Domain-OOS dataset, we can find that the performances become much worse on the larger fine-grained BANKING77-OOS dataset. Especially the in-scope accuracy and OOS precision are pretty low, even with more training examples. This finding encourages more attention to be put on fine-grained intent detection with OOS examples. 



\subsection{Analysis and Discussions}
One key to the OOS detection is a clear separation between in-scope and OOS examples in terms of the model confidence score~\cite{zhang2020discriminative}.
Figure~\ref{fig:model-comp} illustrates the differences in confidence score distributions.
The confidence scores of ID-OOS examples are close or mixed with the scores of in-scope intents, and are higher than the OOD-OOS examples, showing that separating ID-OOS examples is much harder than separating OOD-OOS examples. 

Among comparisons of the pre-trained models, ALBERT performs worst, and  RoBERTa performs better than other models in general since the confidence score received by in-scope examples is higher than that received by the OOS examples.  Figure~\ref{fig:comparisons-among-models} also shows similar results. We conjecture that pre-trained models with more data, better architecture and objectives, etc., are relatively more robust to OOD-OOS and ID-OOS examples than the others. 
Comparing the RoBERTa 5-shot and full-shot confidence distributions, the ID-OOS confidence scores are improved, indicating over-confidence to separate semantically-related intents (i.e., ID-OOS examples).

Next, we inspect what ID-OOS examples are misclassified, and we take RoBERTa as an example as it performs better than other models in general.  
Figure~\ref{fig:conf_mat} shows the confusion matrices of RoBERTa w.r.t. the ``Banking'' domain in the CLINC-Single-Domain-OOS dataset, under full-shot setting.
We can see that the model is extremely likely to confuse ID-OOS intents with particular in-scope intents.
We expect this is from our ID-OOS design, and the trend is consistent across evaluated models.

Now one question arises: \textit{what causes the model's mistakes?}
One presumable source is the keyword overlap. We checked unigram overlap, after removing stop words, for the intent pairs with the three darkest colors in ``Banking'' based on Figure~\ref{fig:conf_mat}.
We then masked top-5 overlapped unigrams from the corresponding intent examples in the development set using the {\it mask} token in the RoBERTa masked language model pretraining and conducted the same evaluation.\footnote{We did not mask the top-10 or top-15 overlapped unigrams, 
as many tokens are already masked in the user utterance when setting the threshold to 5, as shown in Table~\ref{tb:word_overlap}.}
Figure~\ref{fig:conf_mat} shows that most of the confusing intent pairs are still misclassified even without the keyword overlap.
Table~\ref{tb:word_overlap} shows two intent pairs with the overlapped words and their masked ID-OOS examples.
It is surprising that the examples show counterintuitive results. That is, even with the aggressive masking, the model still tends to assign high confidence scores to some other in-scope intents. We also adopted state-of-the-art methods with contrastive learning on few-shot text classification~\citep{liu2021fast} and intent detection~\citep{zhang2021few}. However, we did not achieve promising improvements on OOD-OOS and ID-OOS detection, and we leave more explorations to future work.




\begin{table}[]
\begin{center}
  \resizebox{\linewidth}{!}{
\begin{tabular}{ll}
\hline
Intent pair            & bill\_due \&     bill\_balance                                                                                                                                                                      \\ \hdashline
Unigram overlap        & bill (60), pay (9), need (9), know (8), due (7)                                                                                                                                                     \\ \hdashline
Masked ID-OOS example & \begin{tabular}[c]{@{}l@{}}i {[}mask{]} to {[}mask{]} what day i {[}mask{]} to {[}mask{]} \\ my water {[}mask{]}  $\rightarrow$ bill\_balance (confidence: \textbf{0.84})\end{tabular} \\ \hline
Intent pair            & improve\_credit\_score \& credit\_score                                                                                                                                                             \\ \hdashline
Unigram overlap        & credit (99), score (76), tell (7), want (3), like (3)                                                                                                                                               \\ \hdashline
Masked ID-OOS example & \begin{tabular}[c]{@{}l@{}}i'd {[}mask{]} to make my {[}mask{]} {[}mask{]} better \\ $\rightarrow$ credit\_limit\_change (confidence: \textbf{0.86})\end{tabular}                     \\ \hline
\end{tabular}}\caption{Examples investigated for the unigram overlap analysis. The overlap frequency is also presented.}\label{tb:word_overlap}
\end{center}
\vspace{-1.3em}
\end{table}

\section{Conclusion}
\label{concolusion}

We have investigated the robustness of pre-trained Transformers in few-shot intent detection with OOS samples.
Our results on two new constructed datasets show that pre-trained models are not robust on ID-OOS examples. Both the OOS detection tasks are challenging in the scenario of fine-grained intent detection.
Our work encourages more attention to be put on the above findings.



\clearpage
\bibliographystyle{acl_natbib}
\bibliography{reference}

\begin{thebibliography}{21}
\expandafter\ifx\csname natexlab\endcsname\relax\def\natexlab#1{#1}\fi

\bibitem[{Casanueva et~al.(2020)Casanueva, Tem{\v{c}}inas, Gerz, Henderson, and
  Vuli{\'c}}]{casanueva2020efficient}
I{\~n}igo Casanueva, Tadas Tem{\v{c}}inas, Daniela Gerz, Matthew Henderson, and
  Ivan Vuli{\'c}. 2020.
\newblock Efficient intent detection with dual sentence encoders.
\newblock In \emph{Proceedings of the 2nd Workshop on Natural Language
  Processing for Conversational AI}, pages 38--45.

\bibitem[{Cavalin et~al.(2020)Cavalin, Ribeiro, Appel, and
  Pinhanez}]{cavalin2020improving}
Paulo Cavalin, Victor Henrique~Alves Ribeiro, Ana Appel, and Claudio Pinhanez.
  2020.
\newblock Improving out-of-scope detection in intent classification by using
  embeddings of the word graph space of the classes.
\newblock In \emph{EMNLP}, pages 3952--3961.

\bibitem[{Clark et~al.(2020)Clark, Luong, Le, and Manning}]{electra}
Kevin Clark, Minh-Thang Luong, Quoc~V. Le, and Christopher~D. Manning. 2020.
\newblock Electra: Pre-training text encoders as discriminators rather than
  generators.
\newblock In \emph{ICLR}.

\bibitem[{Devlin et~al.(2019)Devlin, Chang, Lee, and Toutanova}]{bert}
Jacob Devlin, Ming-Wei Chang, Kenton Lee, and Kristina Toutanova. 2019.
\newblock {{BERT}: Pre-training of Deep Bidirectional Transformers for Language
  Understanding}.
\newblock In \emph{NAACL-HLT}, pages 4171--4186.

\bibitem[{Hendrycks and Gimpel(2017)}]{softmax_conf}
Dan Hendrycks and Kevin Gimpel. 2017.
\newblock A baseline for detecting misclassified and out-of-distribution
  examples in neural networks.
\newblock In \emph{ICLR}.

\bibitem[{Hendrycks et~al.(2020{\natexlab{a}})Hendrycks, Liu, Wallace,
  Dziedzic, Krishnan, and Song}]{acl-ood}
Dan Hendrycks, Xiaoyuan Liu, Eric Wallace, Adam Dziedzic, Rishabh Krishnan, and
  Dawn Song. 2020{\natexlab{a}}.
\newblock {Pretrained Transformers Improve Out-of-Distribution Robustness}.
\newblock In \emph{ACL}, pages 2744--2751.

\bibitem[{Hendrycks et~al.(2020{\natexlab{b}})Hendrycks, Liu, Wallace,
  Dziedzic, Krishnan, and Song}]{adamwr}
Dan Hendrycks, Xiaoyuan Liu, Eric Wallace, Adam Dziedzic, Rishabh Krishnan, and
  Dawn Song. 2020{\natexlab{b}}.
\newblock {Pretrained Transformers Improve Out-of-Distribution Robustness}.
\newblock \emph{arXiv preprint arXiv:2004.06100}.

\bibitem[{Lan et~al.(2020)Lan, Chen, Goodman, Gimpel, Sharma, and
  Soricut}]{albert}
Zhenzhong Lan, Mingda Chen, Sebastian Goodman, Kevin Gimpel, Piyush Sharma, and
  Radu Soricut. 2020.
\newblock {ALBERT: A Lite BERT for Self-supervised Learning of Language
  Representations}.
\newblock In \emph{ICLR}.

\bibitem[{Larson et~al.(2019)Larson, Mahendran, Peper, Clarke, Lee, Hill,
  Kummerfeld, Leach, Laurenzano, Tang, and Mars}]{oos-intent}
Stefan Larson, Anish Mahendran, Joseph~J. Peper, Christopher Clarke, Andrew
  Lee, Parker Hill, Jonathan~K. Kummerfeld, Kevin Leach, Michael~A. Laurenzano,
  Lingjia Tang, and Jason Mars. 2019.
\newblock {An Evaluation Dataset for Intent Classification and Out-of-Scope
  Prediction}.
\newblock In \emph{EMNLP}, pages 1311--1316.

\bibitem[{Liu et~al.(2021)Liu, Vuli{\'c}, Korhonen, and Collier}]{liu2021fast}
Fangyu Liu, Ivan Vuli{\'c}, Anna Korhonen, and Nigel Collier. 2021.
\newblock Fast, effective, and self-supervised: Transforming masked language
  models into universal lexical and sentence encoders.
\newblock In \emph{EMNLP}, pages 1442--1459.

\bibitem[{Liu et~al.(2019)Liu, Ott, Goyal, Du, Joshi, Chen, Levy, Lewis,
  Zettlemoyer, and Stoyanov}]{roberta}
Yinhan Liu, Myle Ott, Naman Goyal, Jingfei Du, Mandar Joshi, Danqi Chen, Omer
  Levy, Mike Lewis, Luke Zettlemoyer, and Veselin Stoyanov. 2019.
\newblock {RoBERTa: A Robustly Optimized BERT Pretraining Approach}.
\newblock \emph{arXiv preprint arXiv:1907.11692}.

\bibitem[{Pascanu et~al.(2013)Pascanu, Mikolov, and Bengio}]{clip}
Razvan Pascanu, Tomas Mikolov, and Yoshua Bengio. 2013.
\newblock {On the difficulty of training recurrent neural networks}.
\newblock In \emph{Proceedings of the 30th International Conference on Machine
  Learning (ICML)}, pages 1310--1318.

\bibitem[{Shu et~al.(2017)Shu, Xu, and Liu}]{open-class}
Lei Shu, Hu~Xu, and Bing Liu. 2017.
\newblock {{DOC}: Deep Open Classification of Text Documents}.
\newblock In \emph{EMNLP}, pages 2911--2916.

\bibitem[{Wolf et~al.(2019)Wolf, Debut, Sanh, Chaumond, Delangue, Moi, Cistac,
  Rault, Louf, Funtowicz et~al.}]{wolf2019huggingface}
Thomas Wolf, Lysandre Debut, Victor Sanh, Julien Chaumond, Clement Delangue,
  Anthony Moi, Pierric Cistac, Tim Rault, R{\'e}mi Louf, Morgan Funtowicz,
  et~al. 2019.
\newblock Huggingface's transformers: State-of-the-art natural language
  processing.
\newblock \emph{arXiv preprint arXiv:1910.03771}.

\bibitem[{Wu et~al.(2020)Wu, Hoi, Socher, and Xiong}]{tod-bert}
Chien-Sheng Wu, Steven Hoi, Richard Socher, and Caiming Xiong. 2020.
\newblock {ToD-BERT: Pre-trained Natural Language Understanding for
  Task-Oriented Dialogues}.
\newblock \emph{EMNLP}.

\bibitem[{Xie et~al.(2022)Xie, Yang, Lin, Wu, Hashimoto, Qu, Kang, Yin, Wang,
  Yavuz et~al.}]{xie2022converse}
Tian Xie, Xinyi Yang, Angela~S Lin, Feihong Wu, Kazuma Hashimoto, Jin Qu,
  Young~Mo Kang, Wenpeng Yin, Huan Wang, Semih Yavuz, et~al. 2022.
\newblock Converse--a tree-based modular task-oriented dialogue system.
\newblock \emph{arXiv preprint arXiv:2203.12187}.

\bibitem[{Xu et~al.(2021)Xu, Ren, Zhang, Feng, and Xiong}]{xu2021unsupervised}
Keyang Xu, Tongzheng Ren, Shikun Zhang, Yihao Feng, and Caiming Xiong. 2021.
\newblock Unsupervised out-of-domain detection via pre-trained transformers.
\newblock In \emph{ACL}, pages 1052--1061.

\bibitem[{Zhan et~al.(2021)Zhan, Liang, Liu, Fan, Wu, and Lam}]{zhan2021out}
Li-Ming Zhan, Haowen Liang, Bo~Liu, Lu~Fan, Xiao-Ming Wu, and Albert~YS Lam.
  2021.
\newblock Out-of-scope intent detection with self-supervision and
  discriminative training.
\newblock In \emph{ACL}, pages 3521--3532.

\bibitem[{Zhang et~al.(2021)Zhang, Bui, Yoon, Chen, Liu, Xia, Tran, Chang, and
  Philip}]{zhang2021few}
Jianguo Zhang, Trung Bui, Seunghyun Yoon, Xiang Chen, Zhiwei Liu, Congying Xia,
  Quan~Hung Tran, Walter Chang, and S~Yu Philip. 2021.
\newblock Few-shot intent detection via contrastive pre-training and
  fine-tuning.
\newblock In \emph{EMNLP}, pages 1906--1912.

\bibitem[{Zhang et~al.(2020)Zhang, Hashimoto, Liu, Wu, Wan, Philip, Socher, and
  Xiong}]{zhang2020discriminative}
Jianguo Zhang, Kazuma Hashimoto, Wenhao Liu, Chien-Sheng Wu, Yao Wan, S~Yu
  Philip, Richard Socher, and Caiming Xiong. 2020.
\newblock Discriminative nearest neighbor few-shot intent detection by
  transferring natural language inference.
\newblock In \emph{EMNLP}, pages 5064--5082.

\bibitem[{Zheng et~al.(2019)Zheng, Chen, and Huang}]{ood_intent}
Yinhe Zheng, Guanyi Chen, and Minlie Huang. 2019.
\newblock {Out-of-domain Detection for Natural Language Understanding in Dialog
  Systems}.
\newblock \emph{arXiv preprint arXiv:1909.03862}.

\end{thebibliography}

\clearpage

\appendix

\section{More Results}
\begin{figure*}[t]
	\begin{center}
    	\includegraphics[width=\linewidth]{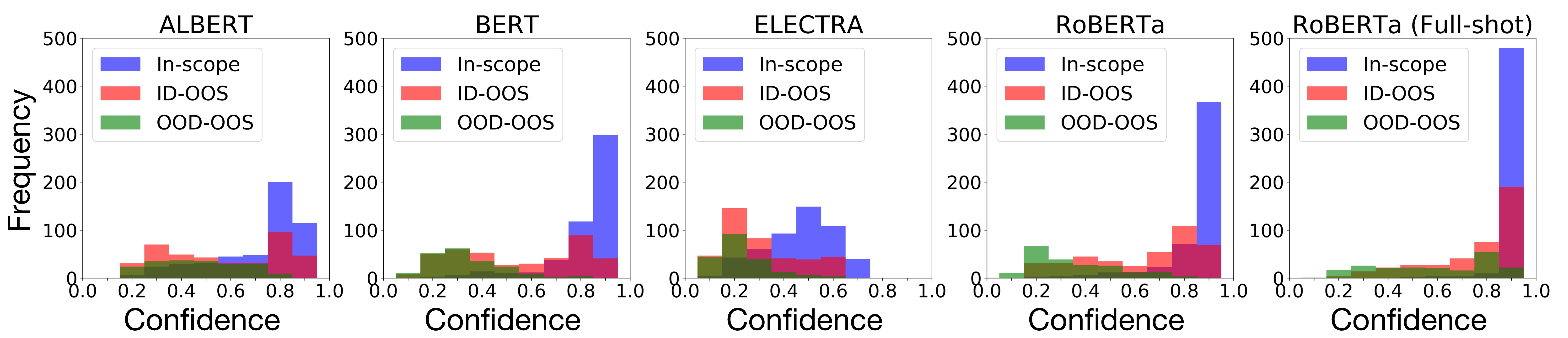}
    \end{center}
\caption{Model confidence on the development set of the ``Credit cards'' domain in CLINC-Single-Domain-OOS dataset under 5-shot setting. Darker colors indicate overlaps.}
\label{fig:model-confidence}
\end{figure*}

\begin{figure*}
	\begin{center}
    	\includegraphics[width=\linewidth]{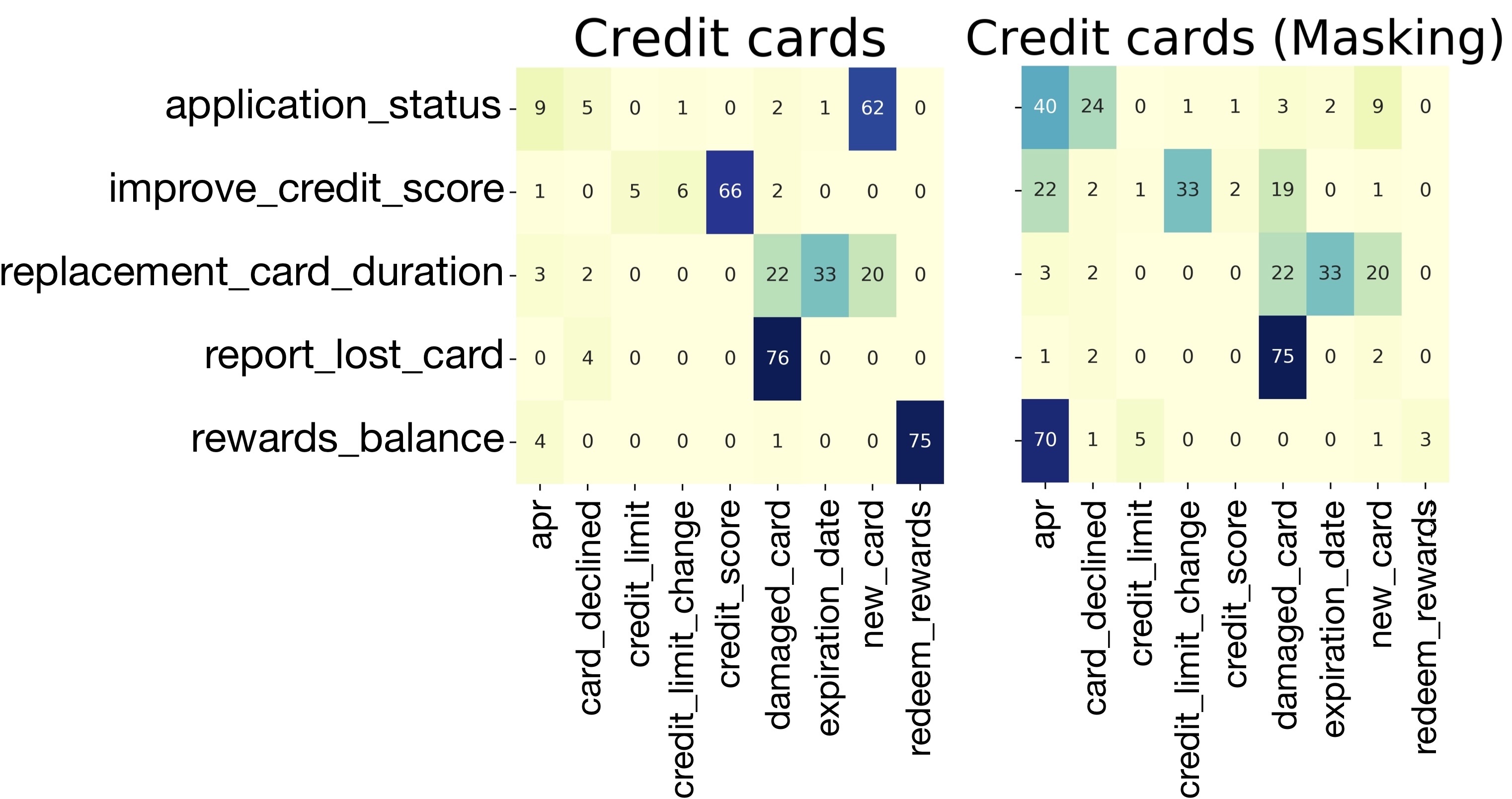}
    \end{center}
\caption{Full-shot confusion matrices on the development set with and without masking (``Credit cards'', RoBERTa). Vertical axis: ID-OOS; horizontal axis: in-scope (only predicted intents considered). 
}
\label{fig:conf_mat_credid_card}
\end{figure*}

\begin{figure*}[ht]
	\begin{center}
    	\includegraphics[width=\linewidth]{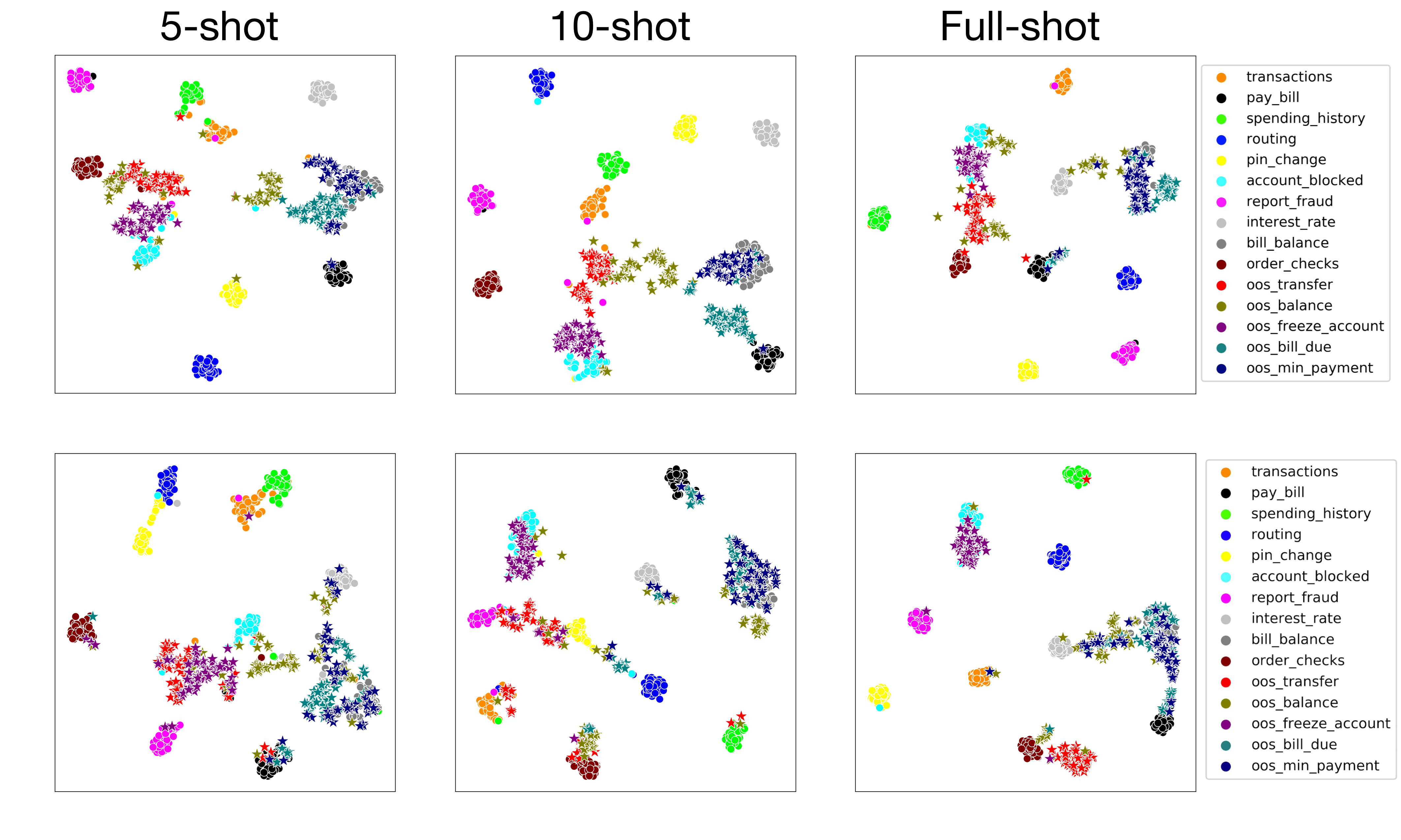}
    \end{center}
\caption{RoBERTa (first row) and ELECTRA (second row) tSNE visualizations on the development set of the ``Banking'' domain in CLINC-Single-Domain-OOS dataset. }
\label{fig:tsne-main}
\end{figure*}

Figure~\ref{fig:model-confidence} shows the model confidence level on the development set of the ``Credit cards'' domain in the CLINC-Single-Domain-OOS dataset. We can see that RoBERTa is relatively more robust with limited data.
Figure~\ref{fig:conf_mat_credid_card} shows the confusion matrices of RoBERTa w.r.t. the ``Credit cards'' domain in the CLINC-Single-Domain-OOS dataset. The model is confused to identify ID-OOS intents.
Figure~\ref{fig:tsne-main} shows the tSNE visualizations for ID-OOS intents w.r.t. the ``Banking'' domain in the CLINC-Single-Domain-OOS dataset.  The models struggle to classify the ID-OOS intents even with more data.

\end{document}